\pdfoutput=1

\documentclass{article}
\usepackage{colm2024_conference}

\usepackage{url}
\usepackage{booktabs}
\usepackage{hyperref}
\usepackage{graphicx}
\usepackage{microtype}
\usepackage{multicol}
\usepackage{xcolor}   
\usepackage{multirow}
\usepackage{lipsum}
\usepackage{enumitem}
\usepackage{pifont}
\usepackage{colortbl}
\usepackage{tabularx}
\usepackage{caption} 
\usepackage{float}
\usepackage{amsmath,amsfonts,amssymb,bbm}

\newcommand{\tabincell}[2]{\begin{tabular}{@{}#1@{}}#2\end{tabular}}
\newcommand*\samethanks[1][\value{footnote}]{\footnotemark[#1]}

\title{\textsc{ProcessBench}: Identifying Process Errors \\ in Mathematical Reasoning}

\author{
\textbf{Chujie Zheng\thanks{Corresponding authors.} \quad Zhenru Zhang \quad Beichen Zhang \quad Runji Lin \quad Keming Lu} \\
\textbf{Bowen Yu\samethanks{} \quad Dayiheng Liu\samethanks{} \quad Jingren Zhou \quad Junyang Lin\samethanks{} } \\
\vspace{0.15cm}
Qwen Team, Alibaba Inc.
}

\begin{document}
\maketitle

\begin{abstract}
As language models regularly make mistakes when solving math problems, automated identification of errors in the reasoning process becomes increasingly significant for their scalable oversight.
In this paper, we introduce \textbf{\textsc{ProcessBench}} for measuring the ability to identify erroneous steps in mathematical reasoning.
It consists of 3,400 test cases, primarily focused on competition- and Olympiad-level math problems.
Each test case contains a step-by-step solution with error location annotated by human experts.
Models are required to identify the earliest step that contains an error, or conclude that all steps are correct.
We conduct extensive evaluation on \textsc{ProcessBench}, involving two types of models: \textit{process reward models (PRMs)} and \textit{critic models}, where for the latter we prompt general language models to critique each solution step by step.
We draw two main observations:
(1) Existing PRMs typically fail to generalize to more challenging math problems beyond GSM8K and MATH.
They underperform both critic models (i.e., prompted general language models) and our own trained PRM that is straightforwardly fine-tuned on the PRM800K dataset.
(2) The best open-source model, QwQ-32B-Preview, has demonstrated the critique capability competitive with the proprietary model GPT-4o, despite that it still lags behind the reasoning-specialized o1-mini.
We hope \textsc{ProcessBench} can foster future research in reasoning process assessment, paving the way toward scalable oversight of language models.

\vspace{10mm}

\begin{figure}[hb]
  \centering
  \includegraphics[width=\linewidth]{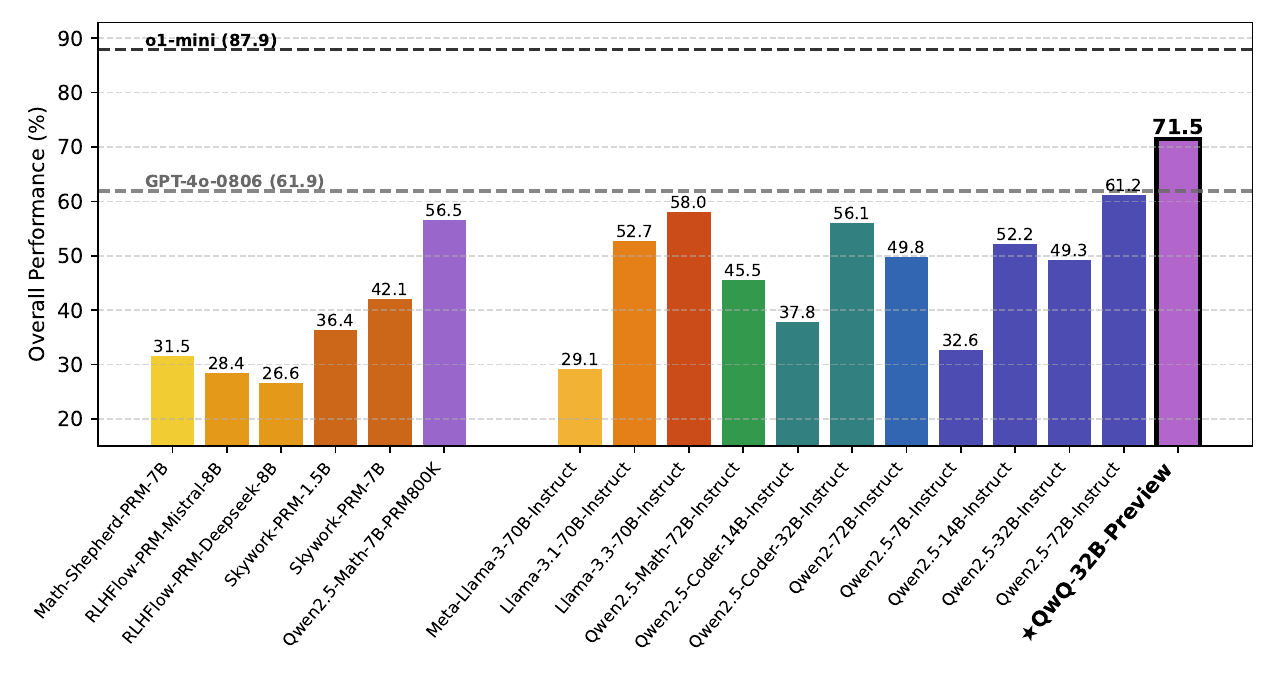}
  \caption{
  Overview of evaluation results on \textsc{ProcessBench} (see Table~\ref{tab:results} for details).
  }
  \label{fig:results_overview}
\end{figure}

\end{abstract}

\section{Introduction}

In recent years, language models have made remarkable progress in complex reasoning tasks, such as mathematics and programming \citep{gpt4o,o1-mini,qwen2,qwen2.5-math,llama3,yi-lightning}, yet they still make mistakes when solving challenging problems.
To achieve scalable oversight \citep{amodei2016concrete,bowman2022measuring,cao2024towards}, i.e., effectively supervising AI systems that get close to or go beyond broadly human-level performance, particularly in complex tasks that are difficult for general humans, we expect language models can identify errors in their reasoning process in an automated way.
However, existing benchmarks related to assessing language models' reasoning process may be hard to satisfy the growing evaluation demand for the error identification ability.
Either their covered problems have become less challenging for recent language models \citep{mathcheck, prm}, or they merely label the correctness of final answers but lack annotations for specific erroneous steps \citep{criticbench}.

\begin{figure}[h]
  \centering
  \vspace{2mm}
  \includegraphics[width=\linewidth]{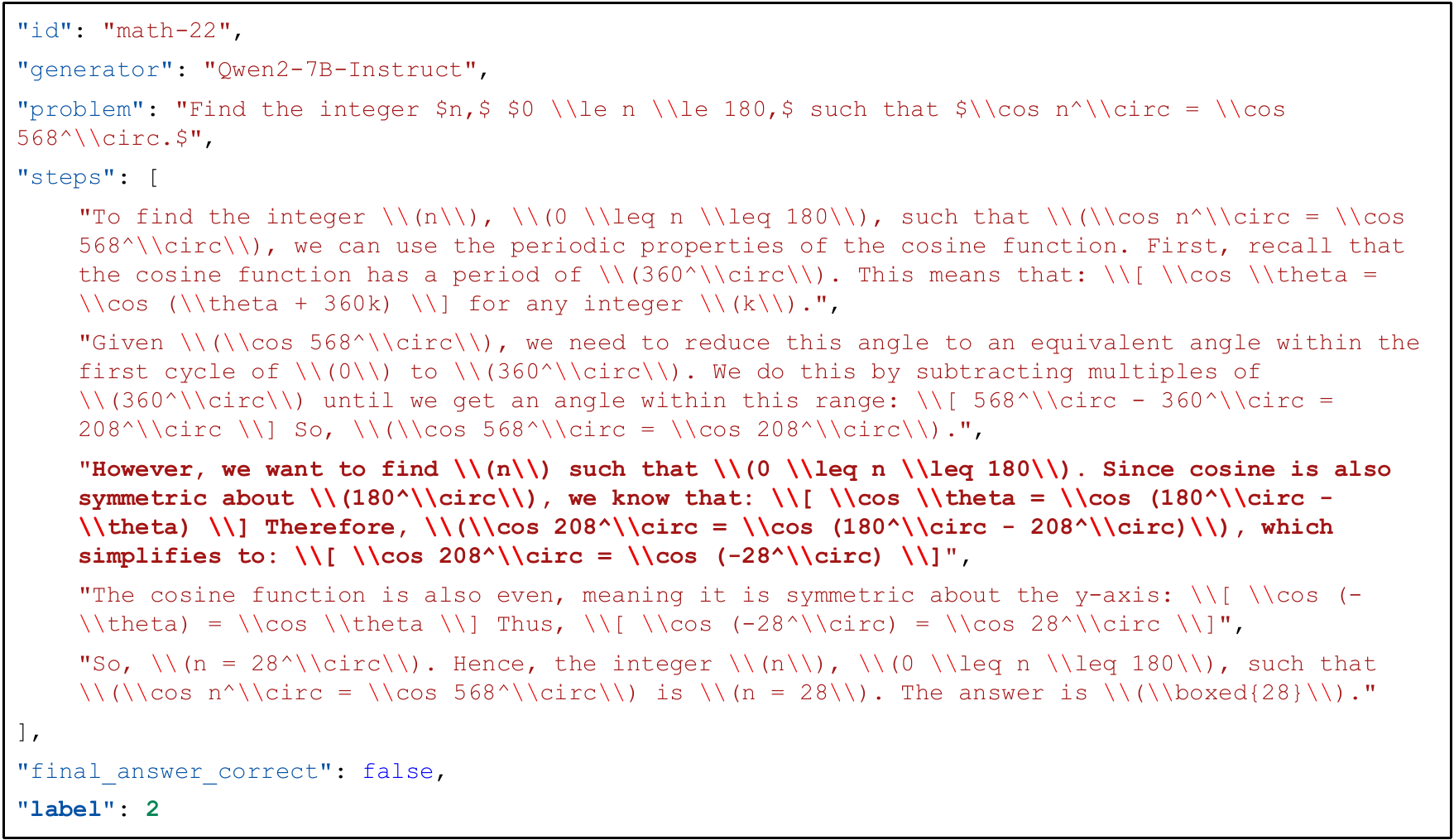}
  \caption{Data example of \textsc{ProcessBench}.
  The label $2$ denotes that the earliest error occurs in the \textbf{2nd step} (indexed from 0).
  For test cases with no errors, the labels are $-1$.
  }
  \vspace{2mm}
  \label{fig:data_example}
\end{figure}

In this paper, we introduce \textbf{\textsc{ProcessBench}} for measuring the ability to identify erroneous steps in mathematical reasoning.
Figure~\ref{fig:data_example} presents a data example.
We prioritize several principles when designing this benchmark:
\begin{itemize}[leftmargin=*]

\item \textbf{Problem difficulty and solution diversity.}
\textsc{ProcessBench} primarily covers competition- and Olympiad-level math problems and utilizes various open-source language models to generate solutions.
This ensures both the difficulty of math problems and the diversity of solution styles, enabling robust evaluation.

\item \textbf{Scale and accuracy.}
\textsc{ProcessBench} consists of 3,400 test cases, with all solutions annotated with error locations by multiple human experts.
The large scale and expert annotation ensure the data quality and the reliability of evaluation.

\item \textbf{Simplicity.}
\textsc{ProcessBench} requires models to identify the earliest erroneous step occurring in the solution, if any exists.
This straightforward evaluation protocol enables easy adaptation for various types of models, such as process reward models (PRMs) and critic models.

\end{itemize}

We conduct extensive evaluation on \textsc{ProcessBench}, involving two types of models: \textit{process reward models (PRMs)} and \textit{critic models}.
For PRMs, we include multiple open-source PRMs \citep{math-shepherd,skywork-prm,rlhflow-prm} to assess the correctness of each reasoning step in the solution.
For critic models, we prompt general language models like Qwen \citep{qwen2,qwen2.5,qwen2.5-coder} and GPT-4o \citep{gpt4o} to critique each solution step by step.
We show that, despite recent growing interest, existing PRMs typically fail to generalize to more challenging math problems beyond GSM8K and MATH.
They underperform both critic models and our own trained PRM that is straightforwardly fine-tuned on the PRM800K dataset, which raises questions about the generalization abilities and scalability of the current data synthesis methodologies used to build PRMs.
In contrast, general language models manifest non-trivial critique capabilities that can not only identify erroneous steps but also provide detailed explanations.
The best open-source model, QwQ-32B-Preview \citep{qwq-32b-preview}, has performed competitively with the proprietary GPT-4o model, while it still lags behind the reasoning-specialized o1-mini \citep{o1-mini}.
We hope \textsc{ProcessBench} can catalyze future research in automated reasoning process assessment, establishing crucial foundations for scalable oversight of language models.

\section{Related Work}

There exist several benchmarks or datasets related to assessing language models' reasoning process.
CriticBench \citep{criticbench} evaluates language models' abilities to critique solutions and correct mistakes in various reasoning tasks.
MathCheck \citep{mathcheck} synthesizes solutions containing erroneous steps using the GSM8K dataset \citep{gsm8k}, in which language models are tasked with judging the correctness of final answers or reasoning steps.
PRM800K \citep{prm} builds on the MATH problems \citep{math} and annotates the correctness and soundness of reasoning steps in model-generated solutions.
It also has sparked a blooming of research interest in building process reward models (PRMs) \citep{math-shepherd,rlhflow-prm,xiong2024building}.

\begin{table}[ht]
  \centering
  \vspace{2mm}
  \caption{Comparison between \textsc{ProcessBench} and other benchmarks or datasets related to reasoning process assessment \citep{criticbench,mathcheck,prm}.
  $^\dagger$: Solution diversity denotes the diversity of language models used for solution generation, corresponding to the ``\# Solution Generators'' column.
  $^\ddagger$: For PRM800K, we only count the 90 complete solutions in its phase 1 test set, as the complete solutions in its phase 2 test set are all terminated at the earliest erroneous steps.
  }
  \vspace{-1mm}
  \scalebox{0.95}{
    \begin{tabular}{lcccccc}
    \toprule
    & \tabincell{c}{\textbf{Problem} \\ \textbf{Diffculty}} & \tabincell{c}{\textbf{\# Solution} \\ \textbf{Generators}} &  \tabincell{c}{\textbf{Solution} \\ \textbf{Diversity}$^\dagger$} & \tabincell{c}{\textbf{Step} \\ \textbf{Annotation?}} & \textbf{Annotator} & \tabincell{c}{\textbf{Test Case Size} \\ \textbf{(Identifying} \\ \textbf{Process Errors)}} \\
    \midrule
    CriticBench & $\bigstar\bigstar$ & 8 & $\bigstar\bigstar\bigstar$ & \ding{55}  & - & - \\
    MathCheck-GSM & $\bigstar$  & 1 & $\bigstar$ & \ding{51}  & Synthetic & 516 \\
    PRM800K & $\bigstar\bigstar$ & 1 & $\bigstar$ & \ding{51}  & Human & 90$^\ddagger$ \\
    \midrule
    \textsc{ProcessBench} & $\bigstar\bigstar\bigstar$ & 12 & $\bigstar\bigstar\bigstar$ & \ding{51} & Human & 3,400 \\
    \bottomrule
    \end{tabular}%
  }
  \vspace{2mm}
  \label{tab:comparison}%
\end{table}%

\textsc{ProcessBench} is distinguished from prior benchmarks or datasets in three key aspects, as highlighted in Table~\ref{tab:comparison}.
\textbf{First}, \textsc{ProcessBench} primarily covers more challenging math problems with competition- or Olympiad-level difficulty, which better fit the rapidly growing capabilities of modern language models.
\textbf{Second}, rather than relying on synthetic data, \textsc{ProcessBench} leverages diverse model-generated natural solutions and employs expert annotation to label erroneous steps, which ensures both real-world applicability and label accuracy.
\textbf{Third}, the large scale of \textsc{ProcessBench} (3,400 test cases in total) enables more comprehensive and robust evaluation.

There has also been extensive research on language models' scalable oversight \citep{amodei2016concrete,bowman2022measuring,cao2024towards} and studies on whether language models can identify the errors in their own outputs.
\citet{prm,math-shepherd,Luo2024OmegaPRM} propose to train specialized reward models to supervise language models' reasoning process (i.e., process reward models or PRMs).
\citet{cannot-self-correct,kamoi-etal-2024-llms} argue that general language models struggle to identify and correct their reasoning errors without external feedback.
\citet{self-critique,criticgpt} show that language models can be trained to write informative critiques for both assisting human evaluation and enabling self-refinement, which favorably scales with increased model capabilities (or model sizes).
We believe the improved capabilities of error identification will build strong foundations for language models' scalable oversight.


\section{Benchmark Construction}

\subsection{Task Definition}
\label{subsec:task_definition}

As shown in Figure~\ref{fig:data_example}, given a math problem and a step-by-step solution, \textsc{ProcessBench} requires models to either identify the \textit{earliest-occurring error}, or conclude that all steps are \textit{correct}.
Formally, given a math problem $P$ and its step-by-step solution $S = \{s_0, ..., s_{n-1}\}$, the task is to output an index $i \in \{-1, 0, ..., n-1\}$, where $i = -1$ indicates that all steps are correct, and $i \geq 0$ indicates that the earliest error occurs at step $s_i$.

Typically but non-inclusively, we consider a step as erroneous if it contains any of the following:
(1) \textbf{Mathematical errors}: incorrect calculations, algebraic manipulations, or formula applications.
(2) \textbf{Logical errors}: invalid deductions, unwarranted assumptions, or flawed reasoning steps.
(3) \textbf{Conceptual errors}: misunderstanding or misapplication of mathematical or problem concepts.
(4) \textbf{Completeness errors}: missing crucial conditions, constraints, or necessary justifications that affect the solution's validity.
Beyond these types of errors, we encourage human annotators to determine the correctness of reasoning steps based on their own expertise.
We do not require human annotators to explicitly annotate error types due to the intractability of intentional categorization.

Note that for steps after the first error, the meaning of their correctness may become ambiguous or debatable: derivations based on incorrect premises can make sense, but still remain on a globally incorrect reasoning path \citep{prm}.
For instance, if step $k$ contains an error in calculating $x = 2$, when it should be $x = 3$, subsequent steps may follow valid algebraic rules but operate on an incorrect value of $x$, making their individual correctness hard to determine.
This is why \textsc{ProcessBench} focuses on identifying the earliest-occurring error in the reasoning process.

\subsection{Data Collection}
\label{subsec:collection}

\paragraph{Problem Curation}
We collect math problems from the test sets of four public and widely used datasets in mathematical reasoning tasks:
GSM8K \citep{gsm8k}, MATH \citep{math}, OlympiadBench \citep{olympiadbench}, and Omni-MATH \citep{omnimath}.
Except for GSM8K, which consists of grade school math problems, the other three datasets all contain problems with competition- or Olympiad-level difficulty.

\paragraph{Solution Generation}
We generate solutions using the widely used Qwen \citep{qwen2,qwen2.5,qwen2.5-math} and LLaMA \citep{llama3} series open-source models, resulting in twelve distinct solution generators in total.
This includes a wide range of model families, sizes, and downstream task performance, leading to the high diversity of solution styles.
Table~\ref{tab:stats_breakdown} in Appendix~\ref{sec:stats_breakdown} presents the breakdown of language models used for \textsc{ProcessBench}'s solution generation.

\paragraph{Solution Reformatting}
In mathematical reasoning tasks, double line breaks (i.e., ``\textbackslash n\textbackslash n'') are commonly used to segment solution steps (or paragraphs).
However, we observed inconsistent step granularity due to varying solution styles and generation randomness.
Some generated solutions frequently used double line breaks, resulting in numerous short, logically incomplete steps, while others used them sparingly, leading to lengthy paragraphs that combine multiple logical components.
Such inconsistency in step granularity (and potential improper step segmentation) would impede the standardization of human annotation criteria.

To address this issue, we adopt a \textit{solution reformatting} method to standardize the step granularity, through which the segmented paragraphs can better correspond to logically complete and progressive reasoning steps.
Specifically, we first replace all the line breaks with white space, and then ask Qwen2.5-72B-Instruct to insert double line breaks (i.e., segment paragraphs) while preserving the solution content.
Since we found that Qwen2.5-72B-Instruct sometimes alters the solution content ($<0.5\%$), we remove those solutions whose final answers change after reformatting (although the content alteration may not influence benchmark construction).
Consequently, the reformatting method effectively unifies the step granularity.
Figure~\ref{fig:reformat} in Appendix~\ref{sec:example_reformat} presents an example of solution reformatting.

\paragraph{Expert Annotation}
To ensure a balance between erroneous and correct solutions, we first use Qwen2.5-72B-Instruct to verify the correctness of final answers in the model-generated solutions against the reference answers.
We then respectively sample solutions with correct or incorrect final answers for subsequent annotation in a balanced way to avoid excessive concentration on solutions from either the weakest or strongest models.

We recruit human experts with doctoral-level mathematical expertise for annotation, and all of them are required to pass the mandatory proficiency examination and annotation tutorial.
The annotators are designated with the same task in \S~\ref{subsec:task_definition}, i.e., identifying the earliest-occurring error in each solution.
However, we notice that the competition- or Olympiad-level math problems can still be challenging even for doctoral students majoring in mathematics.
According to the feedback from the annotators, although they were not required to solve problems from scratch but rather to identify erroneous steps in presented solutions, they would still become quite hesitant in their annotations when uncertain about the correct solution approach, which affected both the annotation speed and quality.
To ease the annotation difficulty, we provide annotators with the reference solutions and answers from the original datasets, while we still explicitly instructed them to inspect and verify the presented model-generated solutions step by step.

Each solution is initially assigned to three different experts.
When the initial three annotators cannot reach consensus, we increase the number of annotators until three of them agree on the same result.
If an agreement cannot be achieved within five annotators (e.g., annotation distribution of $(2,2,1)$ or $(2,1,1,1)$), we discard this solution.
This leads to an overall $\sim 30\%$ discard rate throughout the entire annotation process.
We also discard the solutions where the final answers are incorrect (according to the reference answers) but the human annotation results are correct.
Although such cases are fairly rare ($< 1\%$), they are mostly concentrated in the OlympiadBench and Omni-MATH problems (i.e., Olympiad-level ones).
The agreement statistics in Table~\ref{tab:stats} further evidence that the more challenging problems usually need more annotators to achieve the annotation agreement, particularly for those samples with incorrect final answers.
These results suggest the inherent challenge of our human annotation task.

\subsection{Statistics}
\label{subsec:statistics}

\begin{table}[!ht]
  \centering
  \caption{Statistics of \textsc{ProcessBench}.
  ``\% Process errors'' denotes the proportion of samples with \textit{erroneous reasoning steps} (i.e., annotated as erroneous) among all the samples with \textit{correct final answers}.
  ``\% $\ge n$ steps'' denotes the proportion of samples whose solutions have $\ge n$ steps (split by double line breaks).
  ``\% 3/$n$ agreement'' denotes the proportion of samples where the three-annotator agreement is achieved within $n$ annotators, so $(\%\ 3/3) + (\%\ 3/4) + (\%\ 3/5) = 100\%$.
  }
  \vspace{-1mm}
  \scalebox{0.95}{
    \begin{tabular}{lcccccccc}
    \toprule
     & \multicolumn{2}{c}{\textbf{GSM8K}} & \multicolumn{2}{c}{\textbf{MATH}} & \multicolumn{2}{c}{\textbf{OlympiadBench}} & \multicolumn{2}{c}{\textbf{Omni-MATH}} \\
    \cmidrule(lr){2-3}\cmidrule(lr){4-5}\cmidrule(lr){6-7}\cmidrule(lr){8-9}
    & \textbf{error} & \textbf{correct} & \textbf{error} & \textbf{correct} & \textbf{error} & \textbf{correct} & \textbf{error} & \textbf{correct} \\
    \midrule
    \multirow{1}[0]{*}{\# Samples} & 207   & 193   & 594   & 406   & 661   & 339   & 759   & 241  \\
    \midrule
    \tabincell{l}{\% Process errors \\ (correct final answers)} & \multicolumn{2}{c}{$\frac{200-193}{200}=3.5\%$} & \multicolumn{2}{c}{$\frac{500-406}{500}=18.8\%$} & \multicolumn{2}{c}{$\frac{500-339}{500}=32.2\%$} & \multicolumn{2}{c}{$\frac{500-241}{500}=51.8\%$} \\
    \midrule
    \# Steps & 5.3   & 5.1   & 6.8   & 6.0   & 8.9   & 8.7   & 8.6   & 7.4  \\
    \% $\ge$ 5 steps & 61.8\% & 57.5\% & 73.6\% & 70.4\% & 92.6\% & 92.3\% & 92.5\% & 81.7\% \\
    \% $\ge$ 10 steps & 3.4\% & 1.6\% & 17.8\% & 8.9\% & 33.9\% & 27.1\% & 29.2\% & 21.6\% \\
    \% $\ge$ 15 steps & 0.5\% & 0.0\% & 3.4\% & 2.0\% & 9.1\% & 8.8\% & 7.5\% & 4.1\% \\
    \midrule
    \% 3/3 agreement & 66.7\% & 95.9\% & 59.4\% & 91.9\% & 52.8\% & 85.0\% & 47.8\% & 80.1\% \\
    \% 3/4 agreement & 21.3\% & 3.6\% & 24.4\% & 4.7\% & 24.1\% & 9.1\% & 25.6\% & 13.7\% \\
    \% 3/5 agreement & 12.1\% & 0.5\% & 16.2\% & 3.4\% & 23.1\% & 5.9\% & 26.6\% & 6.2\% \\
    \bottomrule
    \end{tabular}%
  }
  \vspace{2mm}
  \label{tab:stats}%
\end{table}%

\begin{figure}[ht]
  \centering
  \vspace{2mm}
  \includegraphics[width=0.65\linewidth]{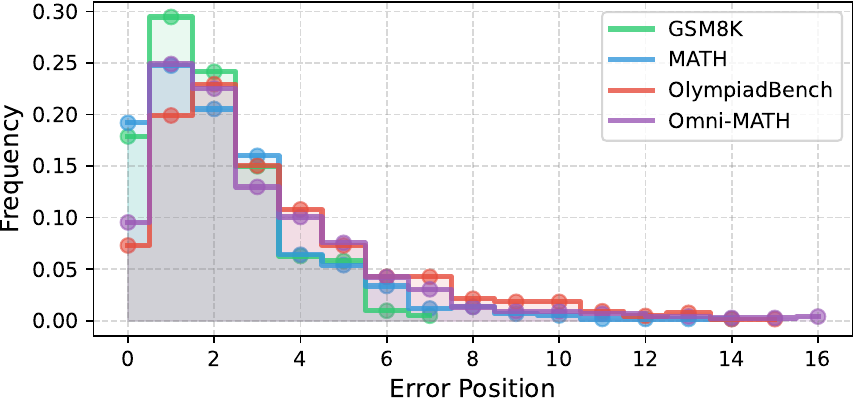}
  \caption{
    Distribution of error positions (indexed from 0; truncated to 16 for better visualization), corresponding to the \textit{label} field as shown in Figure~\ref{fig:data_example}.
  }
  \vspace{2mm}
  \label{fig:error_position_dist}
\end{figure}

The resulting \textsc{ProcessBench} has four subsets, consisting of 3,400 test cases in total.
The detailed statistics are shown in Table~\ref{tab:stats} and Table~\ref{tab:stats_breakdown} (in Appendix~\ref{sec:stats_breakdown}), and we also plot in Figure~\ref{fig:error_position_dist} the distribution of error positions in erroneous samples.
In general, the more challenging the problems, the more solution steps the models generate, and incorrect solutions usually contain more steps than correct ones.
However, across all four subsets, a large proportion of errors occur in the earlier steps, such as steps 0-3 in GSM8K and MATH, and steps 1-5 in OlympiadBench and Omni-MATH.

\begin{figure}[ht]
  \centering
  \includegraphics[width=\linewidth]{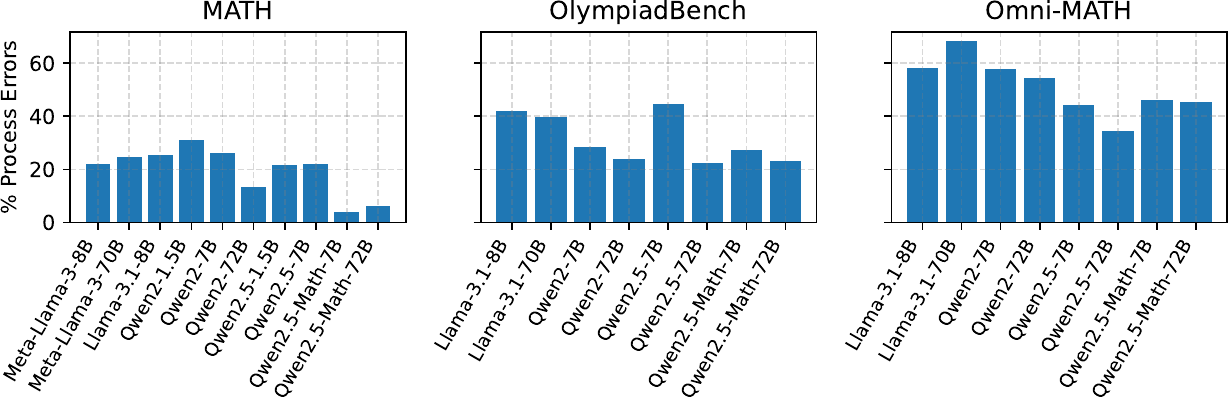}
  \caption{
    Process error ratios per models and subsets, computed as the proportions of samples annotated as \textit{erroneous} among all the samples with \textit{correct final answers} (same as in Table~\ref{tab:stats}).
    The models used for solution generation slightly vary across different subsets, see Table~\ref{tab:stats_breakdown} in Appendix~\ref{sec:stats_breakdown}.
    We observe that no particular models have notably higher process error rates, while \textbf{the process error rates are consistently higher on more difficult problems for all the models.}
  }
  \label{fig:process_error_dist}
\end{figure}

It is noteworthy that \textbf{while we have intentionally controlled an equal number of solutions with incorrect and correct final answers (200 each for GSM8K and 500 each for other subsets)}, the annotation results reveal quite different numbers.
Specifically, in the more challenging subsets like OlympiadBench and Omni-MATH, a larger proportion of solutions with correct final answers still contain erroneous steps.
For instance, in OlympiadBench, $\frac{500-339}{500}=32.2\%$ of solutions with correct final answers are found to contain process errors, while in Omni-MATH this proportion is even higher ($\frac{500-241}{500}=51.8\%$).
In contrast, these proportions in GSM8K and MATH are $\frac{200-193}{200}=3.5\%$ and $\frac{500-406}{500}=18.8\%$, respectively.
In Figure~\ref{fig:process_error_dist}, for each model used for solution generation, we plot the ratio of samples with \textit{erroneous reasoning steps} (i.e., annotated as erroneous) among all the samples with \textit{correct final answers}.
We observe that the process error rates are consistently higher on more difficult problems.
To our knowledge, our work is \textbf{the first to present evidence that on more challenging math problems, current language models are more prone to making process errors even when reaching correct final answers}.
This also suggests the underlying limitation of rule-based RL in mathematical reasoning (i.e., rewarding merely according to the correctness of final answers) and further highlights the significance of identifying errors in the reasoning process.

\section{Evaluation}
\label{sec:evaluation}

\subsection{Setup}
\label{subsec:setup}

For each subset of \textsc{ProcessBench}, we calculate the accuracies on erroneous and correct samples, respectively, and additionally compute their harmonic mean as the \textbf{F1 score}.
We primarily refer to F1 scores to compare model performance, as it balances model behaviors between being overly critical and being incapable of identifying errors.

We consider two types of models in the evaluation on \textsc{ProcessBench}: \textit{process reward models (PRMs)} and \textit{critic models}.

\paragraph{Process Reward Models (PRMs)}
As a recently focal topic, PRMs are proposed to assess and supervise the intermediate steps in language models' reasoning process \citep{prm}, thus naturally falling in the scope of our research.
In practice, PRMs are typically trained using the process labels for intermediate reasoning steps, outputting either the correctness prediction or a scalar score for each reasoning step during inference.
Previous research usually evaluates PRMs based on their improvement in the Best-of-N (BoN) performance of another language model that generates solutions.
However, this lacks a finer-grained inspection on their process assessment abilities, and the evaluation reliability can be heavily affected by the underlying solution generation model.

Our evaluation includes several open-source PRMs: 
(1) Math-Shepherd \citep{math-shepherd}, which obtains the process label for each step via estimating the empirical probability of this step leading to the correct final answer.
(2) Two LLaMA-3.1-based PRMs from \citet{rlhflow-prm}, which roughly follow the training methodology of Math-Shepherd but differ in the solution generation models and optimization objectives.
(3) Two Qwen2.5-Math-based PRMs recently released by \citet{skywork-prm}.
(4) We also train a PRM by fine-tuning Qwen2.5-Math-7B-Instruct on the PRM800K dataset, namely Qwen2.5-Math-7B-PRM800K.
See Appendix~\ref{sec:prm_details} for its training details.

For the (1)(2)(4) PRMs, we extract the earliest erroneous step from their correctness predictions for reasoning steps.
For the (3) PRMs, which produce scalar scores for each reasoning step, we first transform these scores into binary correctness predictions (using a threshold above which steps are considered as correct), and then extract the earliest erroneous step as we do for (1)(2)(4).
The transformation threshold is determined as the one giving the highest F1 score on the GSM8K subset.

\paragraph{Critic Models}
Critic models aim to provide feedback and critique to model-generated texts, non-inclusively including verification, reflection, and correction or refinement.
They have demonstrated promising utility in achieving scalable oversight \citep{self-critique,criticgpt}.

Training critic models for specific domains typically requires significant and specialized effort, which is out of the scope of our work.
Instead, we are more interested in the critique capabilities of \textit{general language models}.
The task definition (\S~\ref{subsec:task_definition}) of \textsc{ProcessBench} enables us to apply simple prompt engineering to repurpose general language models as critic models.
We show in Figure~\ref{fig:prompt} in Appendix~\ref{sec:prompt} the prompt template we implement for our evaluation.
Specifically, models are prompted to return the index of the paragraph where the earliest error occurs as the \textit{final answer}, similar to the conventional evaluation protocol for mathematical reasoning tasks \citep{gsm8k,math,qwen2.5-math}.

Our evaluation includes the widely-used Qwen2 \citep{qwen2}, Qwen2.5 \citep{qwen2.5}, Qwen2.5-Math \citep{qwen2.5-math}, Qwen2.5-Coder \citep{qwen2.5-coder}, and LLaMA-3 \citep{llama3} series open-source models, as well as the recently released QwQ-32B-Preview reasoning model \citep{qwq-32b-preview}.
We also evaluate the proprietary GPT-4o \citep{gpt4o} and o1-mini \citep{o1-mini} models.
We report the performance of open-source models under majority voting over eight samplings, while we also report their performance under greedy decoding in Table~\ref{tab:results_greedy} in Appendix~\ref{sec:supplementary}.
For the proprietary model GPT-4o, we report the results under greedy decoding, while for o1-mini, we report the results under single sampling as its API does not support customized decoding parameters.

\begin{table}[!ht]
  \centering
  \vspace{2mm}
  \caption{
  Evaluation results on \textsc{ProcessBench}.
  We report the F1 score of the respective accuracies on erroneous and correct samples.
  See Table~\ref{tab:results_breakdown} and Table~\ref{tab:results1_breakdown} for breakdown of evaluation results.
  }
  \vspace{-1mm}
  \scalebox{0.95}{
    \begin{tabular}{lccccc}
    \toprule
    \textbf{Model} & \textbf{GSM8K} & \textbf{MATH} & \tabincell{c}{\textbf{Olympiad-} \\ \textbf{Bench}} & \tabincell{c}{\textbf{Omni-} \\ \textbf{MATH}} & \textbf{Average} \\
    \midrule
    \multicolumn{6}{c}{\textit{Open-source \textbf{Process Reward Models (PRMs)}}} \\
    \midrule
    \rowcolor[rgb]{ .988,  .949,  .8} Math-Shepherd-PRM-7B & 47.9  & 29.5  & 24.8  & 23.8  & 31.5  \\
    \rowcolor[rgb]{ .988,  .922,  .8} RLHFlow-PRM-Mistral-8B & 50.4  & 33.4  & 13.8  & 15.8  & 28.4  \\
    \rowcolor[rgb]{ .988,  .922,  .8} RLHFlow-PRM-Deepseek-8B & 38.8  & 33.8  & 16.9  & 16.9  & 26.6  \\
    \rowcolor[rgb]{ .988,  .89,  .8} Skywork-PRM-1.5B & 59.0  & 48.0  & 19.3  & 19.2  & 36.4  \\
    \rowcolor[rgb]{ .988,  .89,  .8} Skywork-PRM-7B & \textbf{70.8} & 53.6  & 22.9  & 21.0  & 42.1  \\
    \rowcolor[rgb]{ .922,  .89,  .988} \textbf{Qwen2.5-Math-7B-PRM800K (our trained)} & 68.2  & \textbf{62.6} & \textbf{50.7} & \textbf{44.3} & \textbf{56.5} \\
    \midrule
    \multicolumn{6}{c}{\textit{Open-source language models, prompted as \textbf{Critic Models}}} \\
    \midrule
    \rowcolor[rgb]{ .988,  .933,  .8} Meta-Llama-3-8B-Instruct & 13.1  & 13.8  & 4.8   & 12.6  & 11.1  \\
    \rowcolor[rgb]{ .988,  .933,  .8} Meta-Llama-3-70B-Instruct & 52.2  & 22.8  & 21.2  & 20.0  & 29.1  \\
    \rowcolor[rgb]{ .988,  .91,  .8} Llama-3.1-8B-Instruct & 10.9  & 5.1   & 2.8   & 1.6   & 5.1  \\
    \rowcolor[rgb]{ .988,  .91,  .8} Llama-3.1-70B-Instruct & 74.9  & 48.2  & 46.7  & 41.0  & 52.7  \\
    \rowcolor[rgb]{ .988,  .882,  .8} Llama-3.3-70B-Instruct & 82.9  & 59.4  & 46.7  & 43.0  & 58.0  \\
    \rowcolor[rgb]{ .882,  .949,  .89} Qwen2.5-Math-7B-Instruct & 26.8  & 25.7  & 14.2  & 12.7  & 19.9  \\
    \rowcolor[rgb]{ .882,  .949,  .89} Qwen2.5-Math-72B-Instruct & 65.8  & 52.1  & 32.5  & 31.7  & 45.5  \\
    \rowcolor[rgb]{ .882,  .933,  .933} Qwen2.5-Coder-7B-Instruct & 14.3  & 6.5   & 4.1   & 1.8   & 6.7  \\
    \rowcolor[rgb]{ .882,  .933,  .933} Qwen2.5-Coder-14B-Instruct & 50.1  & 39.9  & 34.0  & 27.3  & 37.8  \\
    \rowcolor[rgb]{ .882,  .933,  .933} Qwen2.5-Coder-32B-Instruct & 68.9  & 60.1  & 48.9  & 46.3  & 56.1  \\
    \rowcolor[rgb]{ .882,  .922,  .969} Qwen2-7B-Instruct & 8.4   & 19.0  & 14.7  & 12.1  & 13.6  \\
    \rowcolor[rgb]{ .882,  .922,  .969} Qwen2-72B-Instruct & 67.6  & 49.2  & 42.1  & 40.2  & 49.8  \\
    \rowcolor[rgb]{ .89,  .89,  .969} Qwen2.5-7B-Instruct & 36.5  & 36.6  & 29.7  & 27.4  & 32.6  \\
    \rowcolor[rgb]{ .89,  .89,  .969} Qwen2.5-14B-Instruct & 69.3  & 53.3  & 45.0  & 41.3  & 52.2  \\
    \rowcolor[rgb]{ .89,  .89,  .969} Qwen2.5-32B-Instruct & 65.6  & 53.1  & 40.0  & 38.3  & 49.3  \\
    \rowcolor[rgb]{ .89,  .89,  .969} Qwen2.5-72B-Instruct & 76.2  & 61.8  & 54.6  & 52.2  & 61.2  \\
    \rowcolor[rgb]{ .949,  .89,  .988} \boldmath{}\textbf{$\bigstar$ QwQ-32B-Preview}\unboldmath{} & \textbf{88.0} & \textbf{78.7} & \textbf{57.8} & \textbf{61.3} & \textbf{71.5} \\
    \midrule
    \multicolumn{6}{c}{\textit{Proprietary language models, prompted as \textbf{Critic Models}}} \\
    \midrule
    \rowcolor[rgb]{ .906,  .902,  .902} GPT-4o-0806 & 79.2  & 63.6  & 51.4  & 53.5  & 61.9  \\
    \rowcolor[rgb]{ .816,  .808,  .808} o1-mini & 93.2  & 88.9  & 87.2  & 82.4  & 87.9  \\
    \bottomrule
    \end{tabular}%
    }
  \label{tab:results}%
\end{table}%

\subsection{Results}

We present the evaluation results in Table~\ref{tab:results}.
Our observations are summarized as follows:

\paragraph{Generalization Across Difficulty}
From GSM8K and MATH to OlympiadBench and Omni-MATH, with the increased difficulty of math problems, we observe a consistent performance decline for all the models, which suggests the common challenge of both PRMs and critic models in generalization abilities.

\begin{figure}[!t]
  \centering
  \includegraphics[width=\linewidth]{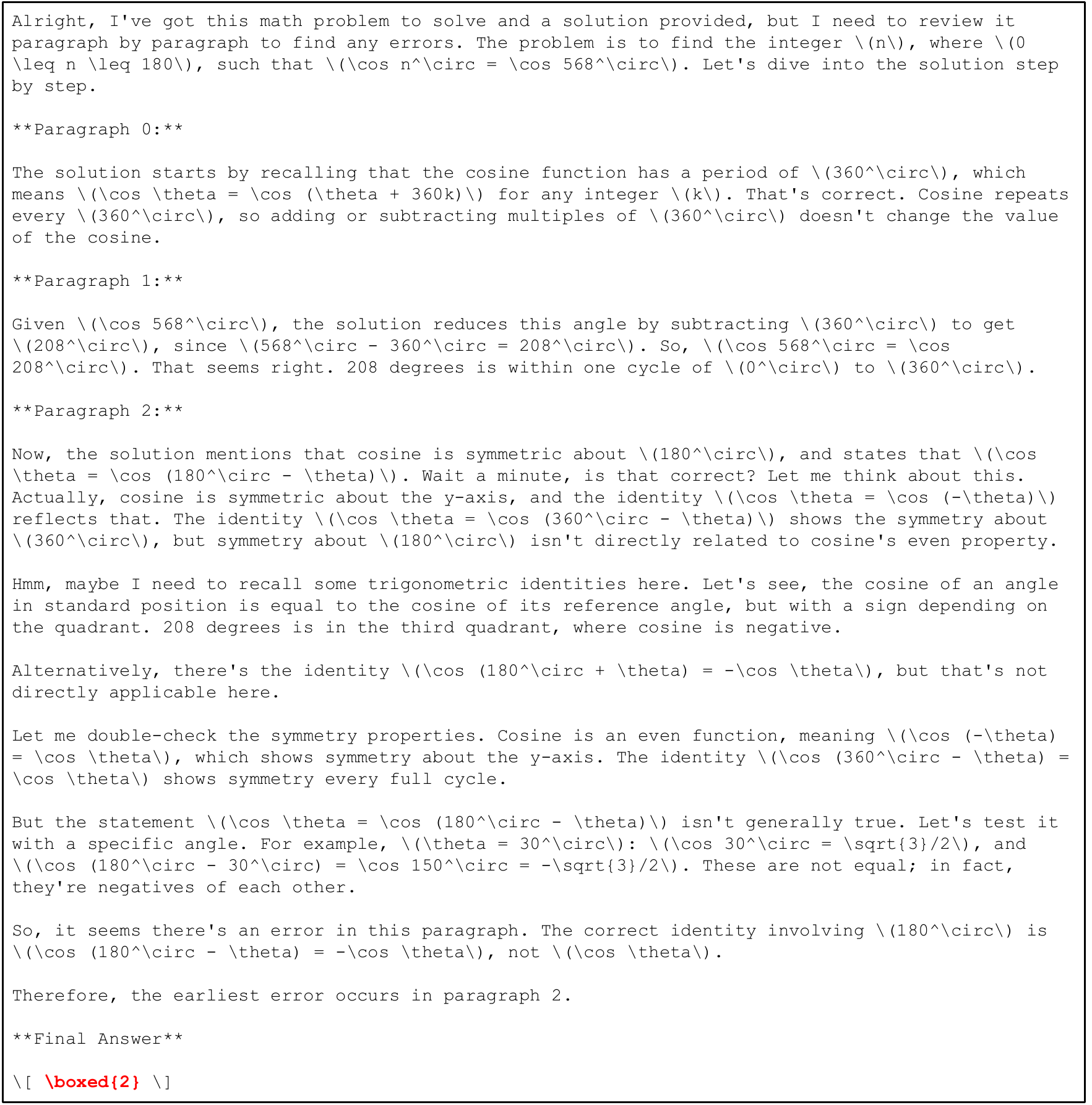}
  \caption{
  Critique generated by QwQ-32B-Preview for the test case in Figure~\ref{fig:data_example}.
  }
  \label{fig:critique}
\end{figure}

\paragraph{Comparison Between PRMs and Critic Models}
We find that existing PRMs typically underperform the top prompt-driven critic models even on the simpler GSM8K and MATH subsets, suggesting that these PRMs struggle to indicate the correctness of the intermediate steps in mathematical reasoning.
Moreover, when moving toward the more challenging OlympiadBench and Omni-MATH subsets, PRMs suffer from a more notable performance decline than critic models.
This raises our concerns about the \textbf{generalization abilities and scalability of the current data synthesis methodologies used to build PRMs}.
More specifically, current methodologies, as exemplified by Math-Shepherd \citep{math-shepherd}, measure the correctness of an intermediate step by estimating the empirical probability of this step leading to the correct final answer.
This kind of approach has two intuitive major issues:
(1) The process labels heavily depend on the language model used to generate solutions (i.e., highly ``on-policy''), which would naturally fail to indicate the correctness of reasoning steps generated by other models.
(2) As demonstrated in \S~\ref{subsec:statistics}, current language models are prone to making process errors even when reaching correct final answers.
This could substantially invalidate the estimated process labels, particularly on the more challenging math problems.
In contrast, \textbf{Qwen2.5-Math-7B-PRM800K}, which is straightforwardly fine-tuned on the \textit{fully human-annotated} PRM800K training set, exhibits significantly stronger performance and generalization ability than other PRMs.

\paragraph{Comparison Among Critic Models}
Compared to PRMs, critic models can benefit from separate reasoning processes when critiquing solutions, as they can ``think'' more before indicating the correctness of each solution step, which leads to their better performance in this error identification task.
Within the same model family, the error identification performance favorably scales with increased model sizes.
Notably, the recently released reasoning model \textbf{\textbf{QwQ-32B-Preview} performs best among the open-source models and is highly competitive with GPT-4o}.
It is noteworthy that QwQ-32B-Preview achieves more \textit{balanced accuracies on erroneous and correct samples} (see Table~\ref{tab:results_breakdown} and~\ref{tab:results1_breakdown} in Appendix~\ref{sec:supplementary}).
We show in Figure~\ref{fig:critique} an example of critique generated by QwQ-32B-Preview to the test case in Figure~\ref{fig:data_example}, which not only identifies the erroneous step but also provides the detailed thinking process and explanation.
Nevertheless, QwQ-32B-Preview still lags behind o1-mini, suggesting that although the gap in problem-solving performance is getting closer between open-source and proprietary models, there still exists another large gap in their critique capabilities.

\section{Conclusion}

We introduce the \textsc{ProcessBench} benchmark for measuring the ability to identify erroneous steps in mathematical reasoning, characterized by its high problem difficulty and solution diversity, large scale, rigorous human annotation, and simple evaluation protocol.
Through extensive evaluation with existing process reward models (PRMs) and prompt-driven critic models, we draw two main observations:
(1) Existing PRMs typically underperform critic models in identifying erroneous reasoning steps, and struggle more to generalize to challenging math problems.
(2) Open-source language models, as exemplified by QwQ-32B-Preview, have demonstrated critique capabilities competitive with the proprietary model GPT-4o, yet still lag behind the reasoning-specialized o1-mini model.
We envision \textsc{ProcessBench} as a cornerstone testbed for advancing automated reasoning process assessment, a critical step toward achieving scalable oversight of language models.

\paragraph{Limitations}
Despite our best efforts throughout the entire benchmark construction process (\S~\ref{subsec:collection}), \textsc{ProcessBench} may still contain inaccurate labels of error locations, particularly for the more challenging Olympiad-level math problems.
Additionally, the solutions discarded in human annotation (\S~\ref{subsec:collection} may involve the particularly challenging problems, which could bias the problem distribution in \textsc{ProcessBench}, although such samples may have exceeded the capabilities of the human annotators in our annotation task.

\bibliographystyle{plainnat}
\bibliography{reference}

\begin{thebibliography}{29}
\providecommand{\natexlab}[1]{#1}
\providecommand{\url}[1]{\texttt{#1}}
\expandafter\ifx\csname urlstyle\endcsname\relax
  \providecommand{\doi}[1]{doi: #1}\else
  \providecommand{\doi}{doi: \begingroup \urlstyle{rm}\Url}\fi

\bibitem[Amodei et~al.(2016)Amodei, Olah, Steinhardt, Christiano, Schulman, and Man{\'e}]{amodei2016concrete}
Dario Amodei, Chris Olah, Jacob Steinhardt, Paul Christiano, John Schulman, and Dan Man{\'e}.
\newblock Concrete problems in ai safety.
\newblock \emph{arXiv preprint arXiv:1606.06565}, 2016.

\bibitem[Bowman et~al.(2022)Bowman, Hyun, Perez, Chen, Pettit, Heiner, Luko{\v{s}}i{\=u}t{\.e}, Askell, Jones, Chen, et~al.]{bowman2022measuring}
Samuel~R Bowman, Jeeyoon Hyun, Ethan Perez, Edwin Chen, Craig Pettit, Scott Heiner, Kamil{\.e} Luko{\v{s}}i{\=u}t{\.e}, Amanda Askell, Andy Jones, Anna Chen, et~al.
\newblock Measuring progress on scalable oversight for large language models.
\newblock \emph{arXiv preprint arXiv:2211.03540}, 2022.

\bibitem[Cao et~al.(2024)Cao, Lu, Lu, Chen, Ren, Xiang, Liu, Lu, He, Han, et~al.]{cao2024towards}
Boxi Cao, Keming Lu, Xinyu Lu, Jiawei Chen, Mengjie Ren, Hao Xiang, Peilin Liu, Yaojie Lu, Ben He, Xianpei Han, et~al.
\newblock Towards scalable automated alignment of llms: A survey.
\newblock \emph{arXiv preprint arXiv:2406.01252}, 2024.

\bibitem[Cobbe et~al.(2021)Cobbe, Kosaraju, Bavarian, Chen, Jun, Kaiser, Plappert, Tworek, Hilton, Nakano, et~al.]{gsm8k}
Karl Cobbe, Vineet Kosaraju, Mohammad Bavarian, Mark Chen, Heewoo Jun, Lukasz Kaiser, Matthias Plappert, Jerry Tworek, Jacob Hilton, Reiichiro Nakano, et~al.
\newblock Training verifiers to solve math word problems.
\newblock \emph{arXiv preprint arXiv:2110.14168}, 2021.

\bibitem[Dubey et~al.(2024)Dubey, Jauhri, Pandey, Kadian, Al-Dahle, Letman, Mathur, Schelten, Yang, Fan, et~al.]{llama3}
Abhimanyu Dubey, Abhinav Jauhri, Abhinav Pandey, Abhishek Kadian, Ahmad Al-Dahle, Aiesha Letman, Akhil Mathur, Alan Schelten, Amy Yang, Angela Fan, et~al.
\newblock The llama 3 herd of models.
\newblock \emph{arXiv preprint arXiv:2407.21783}, 2024.

\bibitem[Gao et~al.(2024)Gao, Song, Yang, Cai, Miao, Dong, Li, Ma, Chen, Xu, et~al.]{omnimath}
Bofei Gao, Feifan Song, Zhe Yang, Zefan Cai, Yibo Miao, Qingxiu Dong, Lei Li, Chenghao Ma, Liang Chen, Runxin Xu, et~al.
\newblock Omni-math: A universal olympiad level mathematic benchmark for large language models.
\newblock \emph{arXiv preprint arXiv:2410.07985}, 2024.

\bibitem[He et~al.(2024)He, Luo, Bai, Hu, Thai, Shen, Hu, Han, Huang, Zhang, et~al.]{olympiadbench}
Chaoqun He, Renjie Luo, Yuzhuo Bai, Shengding Hu, Zhen~Leng Thai, Junhao Shen, Jinyi Hu, Xu~Han, Yujie Huang, Yuxiang Zhang, et~al.
\newblock Olympiadbench: A challenging benchmark for promoting agi with olympiad-level bilingual multimodal scientific problems.
\newblock \emph{arXiv preprint arXiv:2402.14008}, 2024.

\bibitem[Hendrycks et~al.(2021)Hendrycks, Burns, Kadavath, Arora, Basart, Tang, Song, and Steinhardt]{math}
Dan Hendrycks, Collin Burns, Saurav Kadavath, Akul Arora, Steven Basart, Eric Tang, Dawn Song, and Jacob Steinhardt.
\newblock Measuring mathematical problem solving with the math dataset.
\newblock \emph{arXiv preprint arXiv:2103.03874}, 2021.

\bibitem[Huang et~al.(2023)Huang, Chen, Mishra, Zheng, Yu, Song, and Zhou]{cannot-self-correct}
Jie Huang, Xinyun Chen, Swaroop Mishra, Huaixiu~Steven Zheng, Adams~Wei Yu, Xinying Song, and Denny Zhou.
\newblock Large language models cannot self-correct reasoning yet.
\newblock \emph{arXiv preprint arXiv:2310.01798}, 2023.

\bibitem[Hui et~al.(2024)Hui, Yang, Cui, Yang, Liu, Zhang, Liu, Zhang, Yu, Lu, et~al.]{qwen2.5-coder}
Binyuan Hui, Jian Yang, Zeyu Cui, Jiaxi Yang, Dayiheng Liu, Lei Zhang, Tianyu Liu, Jiajun Zhang, Bowen Yu, Keming Lu, et~al.
\newblock Qwen2.5-coder technical report.
\newblock \emph{arXiv preprint arXiv:2409.12186}, 2024.

\bibitem[Hurst et~al.(2024)Hurst, Lerer, Goucher, Perelman, Ramesh, Clark, Ostrow, Welihinda, Hayes, Radford, et~al.]{gpt4o}
Aaron Hurst, Adam Lerer, Adam~P Goucher, Adam Perelman, Aditya Ramesh, Aidan Clark, AJ~Ostrow, Akila Welihinda, Alan Hayes, Alec Radford, et~al.
\newblock Gpt-4o system card.
\newblock \emph{arXiv preprint arXiv:2410.21276}, 2024.

\bibitem[Kamoi et~al.(2024)Kamoi, Zhang, Zhang, Han, and Zhang]{kamoi-etal-2024-llms}
Ryo Kamoi, Yusen Zhang, Nan Zhang, Jiawei Han, and Rui Zhang.
\newblock When can {LLM}s actually correct their own mistakes? a critical survey of self-correction of {LLM}s.
\newblock \emph{Transactions of the Association for Computational Linguistics}, 12:\penalty0 1417--1440, 2024.
\newblock \doi{10.1162/tacl_a_00713}.
\newblock URL \url{https://aclanthology.org/2024.tacl-1.78}.

\bibitem[Kwon et~al.(2023)Kwon, Li, Zhuang, Sheng, Zheng, Yu, Gonzalez, Zhang, and Stoica]{vllm}
Woosuk Kwon, Zhuohan Li, Siyuan Zhuang, Ying Sheng, Lianmin Zheng, Cody~Hao Yu, Joseph Gonzalez, Hao Zhang, and Ion Stoica.
\newblock Efficient memory management for large language model serving with pagedattention.
\newblock In \emph{Proceedings of the 29th Symposium on Operating Systems Principles}, pages 611--626, 2023.

\bibitem[Lightman et~al.(2023)Lightman, Kosaraju, Burda, Edwards, Baker, Lee, Leike, Schulman, Sutskever, and Cobbe]{prm}
Hunter Lightman, Vineet Kosaraju, Yura Burda, Harri Edwards, Bowen Baker, Teddy Lee, Jan Leike, John Schulman, Ilya Sutskever, and Karl Cobbe.
\newblock Let's verify step by step.
\newblock \emph{arXiv preprint arXiv:2305.20050}, 2023.

\bibitem[Lin et~al.(2024)Lin, Gou, Liang, Luo, Liu, and Yang]{criticbench}
Zicheng Lin, Zhibin Gou, Tian Liang, Ruilin Luo, Haowei Liu, and Yujiu Yang.
\newblock Criticbench: Benchmarking llms for critique-correct reasoning.
\newblock \emph{arXiv preprint arXiv:2402.14809}, 2024.

\bibitem[Luo et~al.(2024)Luo, Liu, Liu, Phatale, Lara, Li, Shu, Zhu, Meng, Sun, and Rastogi]{Luo2024OmegaPRM}
Liangchen Luo, Yinxiao Liu, Rosanne Liu, Samrat Phatale, Harsh Lara, Yunxuan Li, Lei Shu, Yun Zhu, Lei Meng, Jiao Sun, and Abhinav Rastogi.
\newblock Improve mathematical reasoning in language models by automated process supervision.
\newblock \emph{arXiv preprint arXiv:2406.06592}, 2024.

\bibitem[McAleese et~al.(2024)McAleese, Pokorny, Uribe, Nitishinskaya, Trebacz, and Leike]{criticgpt}
Nat McAleese, Rai~Michael Pokorny, Juan Felipe~Ceron Uribe, Evgenia Nitishinskaya, Maja Trebacz, and Jan Leike.
\newblock Llm critics help catch llm bugs.
\newblock \emph{arXiv preprint arXiv:2407.00215}, 2024.

\bibitem[OpenAI(2024)]{o1-mini}
OpenAI.
\newblock Openai o1-mini: Advancing cost-efficient reasoning, 2024.
\newblock URL \url{https://openai.com/index/openai-o1-mini-advancing-cost-efficient-reasoning/}.

\bibitem[Qwen(2024{\natexlab{a}})]{qwen2.5}
Team Qwen.
\newblock Qwen2.5: A party of foundation models, September 2024{\natexlab{a}}.
\newblock URL \url{https://qwenlm.github.io/blog/qwen2.5/}.

\bibitem[Qwen(2024{\natexlab{b}})]{qwq-32b-preview}
Team Qwen.
\newblock Qwq: Reflect deeply on the boundaries of the unknown, November 2024{\natexlab{b}}.
\newblock URL \url{https://qwenlm.github.io/blog/qwq-32b-preview/}.

\bibitem[Saunders et~al.(2022)Saunders, Yeh, Wu, Bills, Ouyang, Ward, and Leike]{self-critique}
William Saunders, Catherine Yeh, Jeff Wu, Steven Bills, Long Ouyang, Jonathan Ward, and Jan Leike.
\newblock Self-critiquing models for assisting human evaluators.
\newblock \emph{arXiv preprint arXiv:2206.05802}, 2022.

\bibitem[Skywork(2024)]{skywork-prm}
o1~Team Skywork.
\newblock Skywork-o1 open series.
\newblock \url{https://huggingface.co/Skywork}, November 2024.
\newblock URL \url{https://huggingface.co/Skywork}.

\bibitem[Wake et~al.(2024)Wake, Wang, Chen, Lv, Li, Huang, Cai, Zheng, Cooper, Dai, et~al.]{yi-lightning}
Alan Wake, Albert Wang, Bei Chen, CX~Lv, Chao Li, Chengen Huang, Chenglin Cai, Chujie Zheng, Daniel Cooper, Ethan Dai, et~al.
\newblock Yi-lightning technical report.
\newblock \emph{arXiv preprint arXiv:2412.01253}, 2024.

\bibitem[Wang et~al.(2024)Wang, Li, Shao, Xu, Dai, Li, Chen, Wu, and Sui]{math-shepherd}
Peiyi Wang, Lei Li, Zhihong Shao, Runxin Xu, Damai Dai, Yifei Li, Deli Chen, Yu~Wu, and Zhifang Sui.
\newblock Math-shepherd: Verify and reinforce {LLM}s step-by-step without human annotations.
\newblock In \emph{Proceedings of the 62nd Annual Meeting of the Association for Computational Linguistics (Volume 1: Long Papers)}, pages 9426--9439, August 2024.
\newblock \doi{10.18653/v1/2024.acl-long.510}.
\newblock URL \url{https://aclanthology.org/2024.acl-long.510}.

\bibitem[Xiong et~al.(2024{\natexlab{a}})Xiong, Shi, Shen, Rosenberg, Qin, Calandriello, Khalman, Joshi, Piot, Saleh, et~al.]{xiong2024building}
Wei Xiong, Chengshuai Shi, Jiaming Shen, Aviv Rosenberg, Zhen Qin, Daniele Calandriello, Misha Khalman, Rishabh Joshi, Bilal Piot, Mohammad Saleh, et~al.
\newblock Building math agents with multi-turn iterative preference learning.
\newblock \emph{arXiv preprint arXiv:2409.02392}, 2024{\natexlab{a}}.

\bibitem[Xiong et~al.(2024{\natexlab{b}})Xiong, Zhang, Jiang, and Zhang]{rlhflow-prm}
Wei Xiong, Hanning Zhang, Nan Jiang, and Tong Zhang.
\newblock An implementation of generative prm.
\newblock \url{https://github.com/RLHFlow/RLHF-Reward-Modeling}, 2024{\natexlab{b}}.

\bibitem[Yang et~al.(2024{\natexlab{a}})Yang, Yang, Hui, Zheng, Yu, Zhou, Li, Li, Liu, Huang, et~al.]{qwen2}
An~Yang, Baosong Yang, Binyuan Hui, Bo~Zheng, Bowen Yu, Chang Zhou, Chengpeng Li, Chengyuan Li, Dayiheng Liu, Fei Huang, et~al.
\newblock Qwen2 technical report.
\newblock \emph{arXiv preprint arXiv:2407.10671}, 2024{\natexlab{a}}.

\bibitem[Yang et~al.(2024{\natexlab{b}})Yang, Zhang, Hui, Gao, Yu, Li, Liu, Tu, Zhou, Lin, et~al.]{qwen2.5-math}
An~Yang, Beichen Zhang, Binyuan Hui, Bofei Gao, Bowen Yu, Chengpeng Li, Dayiheng Liu, Jianhong Tu, Jingren Zhou, Junyang Lin, et~al.
\newblock Qwen2.5-math technical report: Toward mathematical expert model via self-improvement.
\newblock \emph{arXiv preprint arXiv:2409.12122}, 2024{\natexlab{b}}.

\bibitem[Zhou et~al.(2024)Zhou, Liu, Ning, Liu, Wang, Wong, Huang, Wang, and Huang]{mathcheck}
Zihao Zhou, Shudong Liu, Maizhen Ning, Wei Liu, Jindong Wang, Derek~F Wong, Xiaowei Huang, Qiufeng Wang, and Kaizhu Huang.
\newblock Is your model really a good math reasoner? evaluating mathematical reasoning with checklist.
\newblock \emph{arXiv preprint arXiv:2407.08733}, 2024.

\end{thebibliography}

\appendix 
\onecolumn

\section{Example of Solution Reformatting}
\label{sec:example_reformat}

\begin{figure}[!ht]
  \centering
  \includegraphics[width=\linewidth]{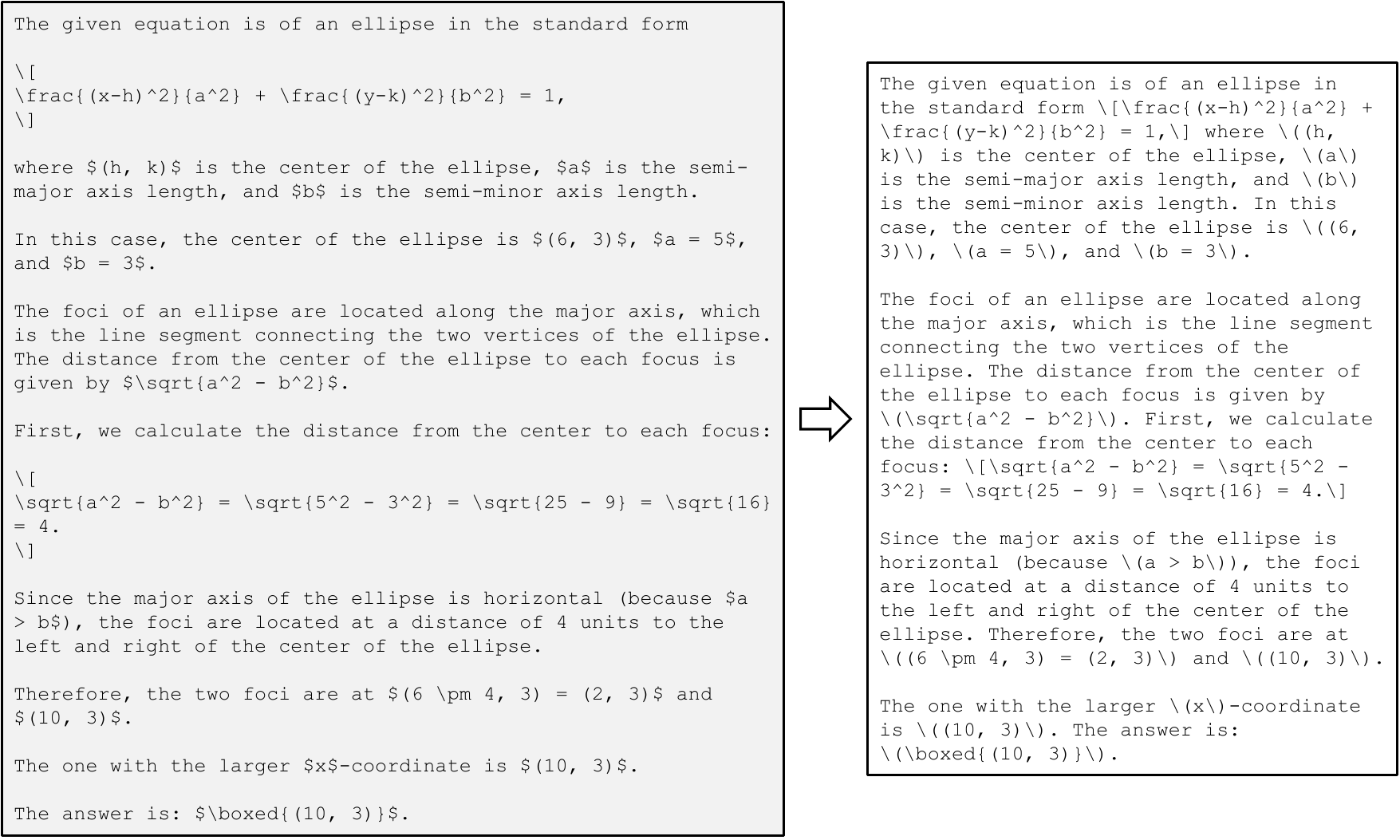}
  \caption{Example of solution reformatting.
  The left is the original solution (generated by Qwen2-7B-Instruct) and the right is the reformatted one.
  The problem, coming from the MATH test set, is ``\textit{The ellipse $\frac{(x-6)^2}{25} + \frac{(y-3)^2}{9} = 1$ has two foci. Find the one with the larger $x$-coordinate. Enter your answer as an ordered pair, like $(2, 1)$}.''
  }
  \label{fig:reformat}
\end{figure}


\section{Breakdown Statistics of \textsc{ProcessBench}}
\label{sec:stats_breakdown}

\begin{table}[!ht]
  \centering
  \caption{Breakdown statistics of \textsc{ProcessBench}.
    $^\dagger$: We encountered a code bug when using Llama-3.1-70B-Instruct and Qwen2.5-72B-Instruct to generate solutions for the MATH problems, thus their counts are all zero in the MATH subset of \textsc{ProcessBench}.
    $^\ddagger$: For the more challenging OlympiadBench and Omni-MATH problems, we exclude models with lower accuracies from subsequent annotation.
    }
  \vspace{-1mm}
    \scalebox{0.95}{
    \begin{tabular}{lcccccccc}
    \toprule
    \multirow{2}[2]{*}{\textbf{Generator}} & \multicolumn{2}{c}{\textbf{GSM8K}} & \multicolumn{2}{c}{\textbf{MATH$^\dagger$}} & \multicolumn{2}{c}{\tabincell{c}{\textbf{OlympiadBench$^\ddagger$}}} & \multicolumn{2}{c}{\tabincell{c}{\textbf{Omni-MATH$^\ddagger$}}} \\
    \cmidrule(lr){2-3}\cmidrule(lr){4-5}\cmidrule(lr){6-7}\cmidrule(lr){8-9}
    & \textbf{error} & \textbf{correct} & \textbf{error} & \textbf{correct} & \textbf{error} & \textbf{correct} & \textbf{error} & \textbf{correct} \\
    \midrule
    Meta-Llama-3-8B-Instruct & 11    & 13    & 56    & 14    & 0     & 0     & 0     & 0  \\
    Meta-Llama-3-70B-Instruct & 16    & 15    & 92    & 49    & 0     & 0     & 0     & 0  \\
    Llama-3.1-8B-Instruct & 38    & 23    & 86    & 53    & 116   & 48    & 131   & 31  \\
    Llama-3.1-70B-Instruct & 7     & 28    & 0     & 0     & 85    & 32    & 103   & 19  \\
    Qwen2-1.5B-Instruct & 37    & 4     & 36    & 11    & 0     & 0     & 0     & 0  \\
    Qwen2-7B-Instruct & 31    & 21    & 89    & 42    & 63    & 45    & 96    & 35  \\
    Qwen2-72B-Instruct & 9     & 11    & 56    & 51    & 64    & 48    & 71    & 25  \\
    Qwen2.5-1.5B-Instruct & 32    & 10    & 31    & 43    & 0     & 0     & 0     & 0  \\
    Qwen2.5-7B-Instruct & 12    & 15    & 62    & 35    & 86    & 37    & 75    & 29  \\
    Qwen2.5-72B-Instruct & 2     & 21    & 0     & 0     & 67    & 38    & 88    & 38  \\
    Qwen2.5-Math-7B-Instruct & 8     & 14    & 47    & 49    & 99    & 48    & 103   & 29  \\
    Qwen2.5-Math-72B-Instruct & 4     & 18    & 39    & 59    & 81    & 43    & 92    & 35  \\
    \midrule
    \multirow{2}[2]{*}{Total} & 207   & 193   & 594   & 406   & 661   & 339   & 759   & 241  \\
    \cmidrule(lr){2-3}\cmidrule(lr){4-5}\cmidrule(lr){6-7}\cmidrule(lr){8-9}
    & \multicolumn{2}{c}{400} & \multicolumn{2}{c}{1,000} & \multicolumn{2}{c}{1,000} & \multicolumn{2}{c}{1,000} \\
    \bottomrule
    \end{tabular}%
    }
  \label{tab:stats_breakdown}%
\end{table}%

\clearpage

\section{Training Details of Qwen2.5-Math-7B-PRM800K}
\label{sec:prm_details}

Qwen2.5-Math-7B-PRM800K is obtained by fine-tuning Qwen2.5-Math-7B-Instruct on the PRM800K training set.
We replace the original language modeling head with a new reward modeling head that outputs binary classification logits.
The classification loss is computed at the second line break positions in all the ``\textbackslash n\textbackslash n''.
We treat the original 1 and 0 labels in PRM800K as our positive labels, while -1 as negative ones.
To eliminate test data contamination, we also remove the PRM800K training samples that have the same problems in \textsc{ProcessBench}.
The training was run on eight A100 80GB GPUs.

\section{Inference Details}

For solution generation in \S~\ref{subsec:collection}, all the models are set with $p=0.9, t=0.7$.
For majority voting evaluation in \S~\ref{sec:evaluation}, we set $p=0.8, t=0.7, k=20$ for Qwen2.5-Math-7/72B-Instruct to ensure their normal generation, while all the other models are set with only $p=0.9$.
All the inference in the evaluation was run with vLLM \citep{vllm} on eight A100 80GB GPUs.

\section{Prompt Template for Critic Model Evaluation}
\label{sec:prompt}

\begin{figure}[!ht]
  \centering
  \includegraphics[width=\linewidth]{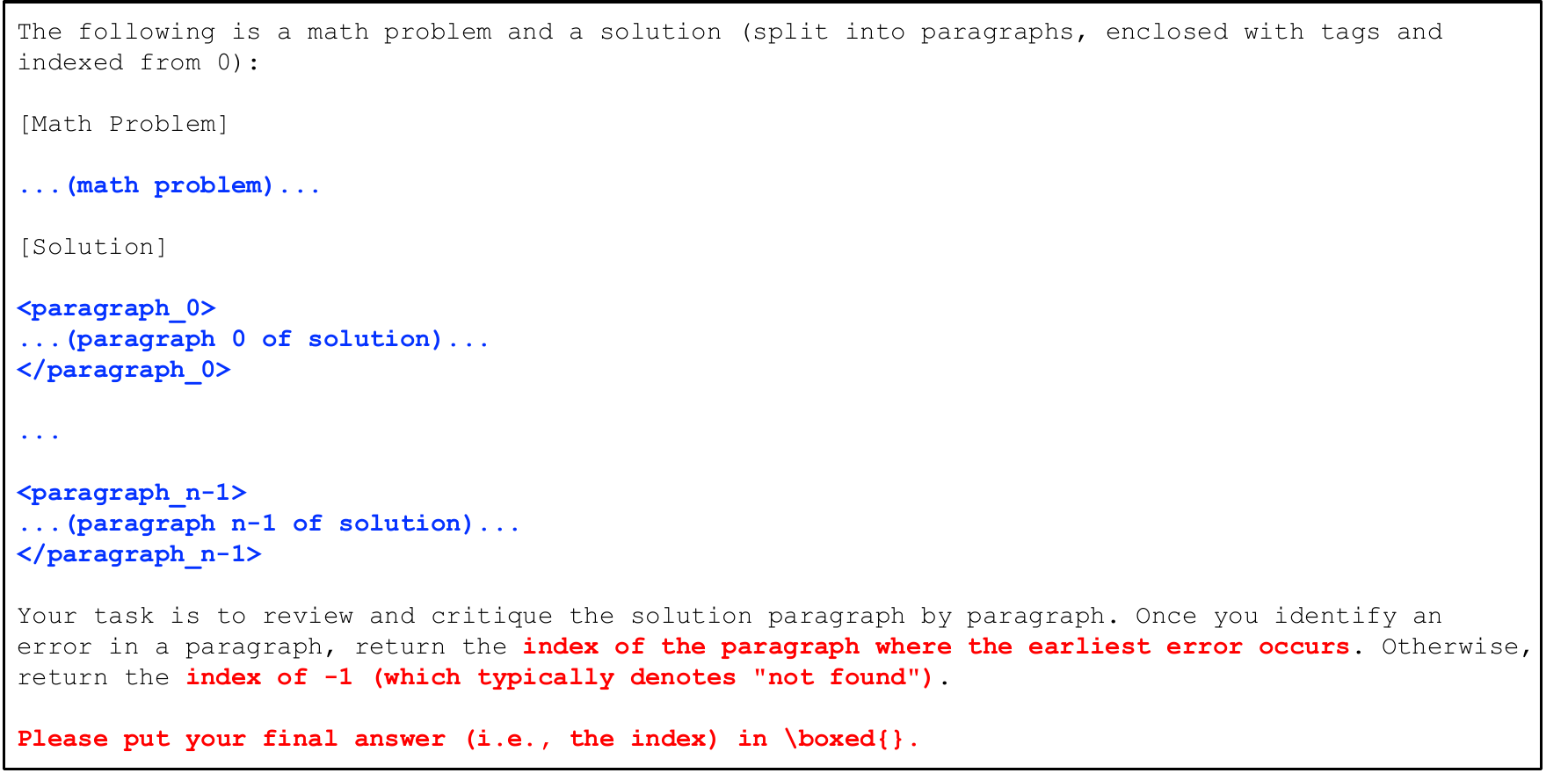}
  \caption{
  Prompt template for critic model evaluation.
  The \textcolor{blue}{\textbf{blue texts}} indicate the input math problem and the solution (split into paragraphs).
  The \textcolor{red}{\textbf{red texts}} describe the required output content and format.
  }
  \label{fig:prompt}
\end{figure}

\clearpage

\section{Supplementary Evaluation Results}
\label{sec:supplementary}

\begin{table}[!ht]
  \centering
  \caption{
  Breakdown of evaluation results on the GSM8K and MATH subsets of \textsc{ProcessBench}.
  The open-source language models  (middle block) are evaluated via \textit{majority voting} over eight samplings.
  }
  \vspace{-1mm}
  \scalebox{0.95}{
    \begin{tabular}{lcccccc}
    \toprule
    \multirow{2}[2]{*}{\textbf{Model}} & \multicolumn{3}{c}{\textbf{GSM8K}} & \multicolumn{3}{c}{\textbf{MATH}} \\
    \cmidrule(lr){2-4}    \cmidrule(lr){5-7}
    & \textbf{error} & \textbf{correct} & \textbf{F1} & \textbf{error} & \textbf{correct} & \textbf{F1} \\
    \midrule
    \multicolumn{7}{c}{\textit{Open-source \textbf{Process Reward Models (PRMs)}}} \\
    \midrule
    \rowcolor[rgb]{ .988,  .949,  .8} Math-Shepherd-PRM-7B & 32.4  & 91.7  & 47.9  & 18.0  & 82.0  & 29.5  \\
    \rowcolor[rgb]{ .988,  .922,  .8} RLHFlow-PRM-Mistral-8B & 33.8  & 99.0  & 50.4  & 21.7  & 72.2  & 33.4  \\
    \rowcolor[rgb]{ .988,  .922,  .8} RLHFlow-PRM-Deepseek-8B & 24.2  & 98.4  & 38.8  & 21.4  & 80.0  & 33.8  \\
    \rowcolor[rgb]{ .988,  .89,  .8} Skywork-PRM-1.5B & 50.2  & 71.5  & 59.0  & 37.9  & 65.3  & 48.0  \\
    \rowcolor[rgb]{ .988,  .89,  .8} Skywork-PRM-7B & 61.8  & 82.9  & \textbf{70.8} & 43.8  & 69.2  & 53.6  \\
    \rowcolor[rgb]{ .922,  .89,  .988} \textbf{Qwen2.5-Math-7B-PRM800K (our trained)} & 53.1  & 95.3  & 68.2  & 48.0  & 90.1  & \textbf{62.6} \\
    \midrule
    \multicolumn{7}{c}{\textit{Open-source language models, prompted as \textbf{Critic Models}}} \\
    \midrule
    \rowcolor[rgb]{ .988,  .933,  .8} Meta-Llama-3-8B-Instruct & 42.5  & 7.8   & 13.1  & 28.6  & 9.1   & 13.8  \\
    \rowcolor[rgb]{ .988,  .933,  .8} Meta-Llama-3-70B-Instruct & 35.7  & 96.9  & 52.2  & 13.0  & 93.3  & 22.8  \\
    \rowcolor[rgb]{ .988,  .91,  .8} Llama-3.1-8B-Instruct & 44.4  & 6.2   & 10.9  & 41.9  & 2.7   & 5.1  \\
    \rowcolor[rgb]{ .988,  .91,  .8} Llama-3.1-70B-Instruct & 64.3  & 89.6  & 74.9  & 35.4  & 75.6  & 48.2  \\
    \rowcolor[rgb]{ .988,  .882,  .8} Llama-3.3-70B-Instruct & 72.5  & 96.9  & 82.9  & 43.3  & 94.6  & 59.4  \\
    \rowcolor[rgb]{ .882,  .949,  .89} Qwen2.5-Math-7B-Instruct & 15.5  & 100.0  & 26.8  & 14.8  & 96.8  & 25.7  \\
    \rowcolor[rgb]{ .882,  .949,  .89} Qwen2.5-Math-72B-Instruct & 49.8  & 96.9  & 65.8  & 36.0  & 94.3  & 52.1  \\
    \rowcolor[rgb]{ .882,  .933,  .933} Qwen2.5-Coder-7B-Instruct & 7.7   & 100.0  & 14.3  & 3.4   & 98.3  & 6.5  \\
    \rowcolor[rgb]{ .882,  .933,  .933} Qwen2.5-Coder-14B-Instruct & 33.8  & 96.4  & 50.1  & 25.4  & 92.4  & 39.9  \\
    \rowcolor[rgb]{ .882,  .933,  .933} Qwen2.5-Coder-32B-Instruct & 54.1  & 94.8  & 68.9  & 44.9  & 90.6  & 60.1  \\
    \rowcolor[rgb]{ .882,  .922,  .969} Qwen2-7B-Instruct & 40.6  & 4.7   & 8.4   & 30.5  & 13.8  & 19.0  \\
    \rowcolor[rgb]{ .882,  .922,  .969} Qwen2-72B-Instruct & 57.0  & 82.9  & 67.6  & 37.7  & 70.9  & 49.2  \\
    \rowcolor[rgb]{ .89,  .89,  .969} Qwen2.5-7B-Instruct & 40.6  & 33.2  & 36.5  & 30.8  & 45.1  & 36.6  \\
    \rowcolor[rgb]{ .89,  .89,  .969} Qwen2.5-14B-Instruct & 54.6  & 94.8  & 69.3  & 38.4  & 87.4  & 53.3  \\
    \rowcolor[rgb]{ .89,  .89,  .969} Qwen2.5-32B-Instruct & 49.3  & 97.9  & 65.6  & 36.7  & 95.8  & 53.1  \\
    \rowcolor[rgb]{ .89,  .89,  .969} Qwen2.5-72B-Instruct & 62.8  & 96.9  & 76.2  & 46.3  & 93.1  & 61.8  \\
    \rowcolor[rgb]{ .949,  .89,  .988} \textbf{QwQ-32B-Preview} & 81.6  & 95.3  & \textbf{88.0} & 78.1  & 79.3  & \textbf{78.7} \\
    \midrule
    \multicolumn{7}{c}{\textit{Proprietary language models, prompted as \textbf{Critic Models}}} \\
    \midrule
    \rowcolor[rgb]{ .906,  .902,  .902} GPT-4o-0806 & 70.0  & 91.2  & 79.2  & 54.4  & 76.6  & 63.6  \\
    \rowcolor[rgb]{ .816,  .808,  .808} o1-mini & 88.9  & 97.9  & 93.2  & 83.5  & 95.1  & 88.9  \\    
    \bottomrule
    \end{tabular}%
  }
  \label{tab:results_breakdown}%
\end{table}%

\begin{table}[!ht]
  \centering
  \caption{
  For the two PRMs from \citet{skywork-prm}, we additionally adjust the threshold (\S~\ref{subsec:setup}) as the one leading to the highest F1 score on each subset (i.e., each subset adopts a respective optimal threshold), which can be viewed as the two PRMs' \textit{upper bound} performance on \textsc{ProcessBench}.
  This table presents the results on the GSM8K and MATH subsets, which are marginally higher than those in Table~\ref{tab:results_breakdown} that all adopt the threshold selected on the GSM8K subset.
  }
  \vspace{-1mm}
  \scalebox{0.95}{
    \begin{tabular}{lcccccc}
    \toprule
    \multirow{2}[2]{*}{\textbf{Model}} & \multicolumn{3}{c}{\textbf{GSM8K}} & \multicolumn{3}{c}{\textbf{MATH}} \\
    \cmidrule(lr){2-4}    \cmidrule(lr){5-7}
    & \textbf{error} & \textbf{correct} & \textbf{F1} & \textbf{error} & \textbf{correct} & \textbf{F1} \\
    \midrule
    \rowcolor[rgb]{ .988,  .89,  .8} Skywork-PRM-1.5B (respective thresholds) & 50.2  & 71.5  & 59.0  & 38.2  & 70.4  & 49.5  \\
    \rowcolor[rgb]{ .988,  .89,  .8} Skywork-PRM-7B (respective thresholds) & 61.8  & 82.9  & 70.8 & 44.1  & 70.9  & 54.4  \\
    \bottomrule
    \end{tabular}%
  }
  \label{tab:results_skywork}%
\end{table}%

\begin{table}[!ht]
  \centering
  \caption{
  Breakdown of evaluation results on the OlympiadBench and Omni-MATH subsets of \textsc{ProcessBench}.
  The open-source language models  (middle block) are evaluated via \textit{majority voting} over eight samplings.
  }
  \vspace{-1mm}
  \scalebox{0.95}{
    \begin{tabular}{lcccccc}
    \toprule
    \multirow{2}[2]{*}{\textbf{Model}} & \multicolumn{3}{c}{\textbf{OlympiadBench}} & \multicolumn{3}{c}{\textbf{Omni-MATH}} \\
    \cmidrule(lr){2-4}    \cmidrule(lr){5-7}
    & \textbf{error} & \textbf{correct} & \textbf{F1} & \textbf{error} & \textbf{correct} & \textbf{F1} \\
    \midrule
    \multicolumn{7}{c}{\textit{Open-source \textbf{Process Reward Models (PRMs)}}} \\
    \midrule
    \rowcolor[rgb]{ .988,  .949,  .8} Math-Shepherd-PRM-7B & 15.0  & 71.1  & 24.8  & 14.2  & 73.0  & 23.8  \\
    \rowcolor[rgb]{ .988,  .922,  .8} RLHFlow-PRM-Mistral-8B & 8.2   & 43.1  & 13.8  & 9.6   & 45.2  & 15.8  \\
    \rowcolor[rgb]{ .988,  .922,  .8} RLHFlow-PRM-Deepseek-8B & 10.1  & 51.0  & 16.9  & 10.1  & 51.9  & 16.9  \\
    \rowcolor[rgb]{ .988,  .89,  .8} Skywork-PRM-1.5B & 15.4  & 26.0  & 19.3  & 13.6  & 32.8  & 19.2  \\
    \rowcolor[rgb]{ .988,  .89,  .8} Skywork-PRM-7B & 17.9  & 31.9  & 22.9  & 14.0  & 41.9  & 21.0  \\
    \rowcolor[rgb]{ .922,  .89,  .988} \textbf{Qwen2.5-Math-7B-PRM800K (our trained)} & 35.7  & 87.3  & \textbf{50.7} & 29.8  & 86.3  & \textbf{44.3} \\
    \midrule
    \multicolumn{7}{c}{\textit{Open-source language models, prompted as \textbf{Critic Models}}} \\
    \midrule
    \rowcolor[rgb]{ .988,  .933,  .8} Meta-Llama-3-8B-Instruct & 27.1  & 2.7   & 4.8   & 26.1  & 8.3   & 12.6  \\
    \rowcolor[rgb]{ .988,  .933,  .8} Meta-Llama-3-70B-Instruct & 12.0  & 92.0  & 21.2  & 11.2  & 91.7  & 20.0  \\
    \rowcolor[rgb]{ .988,  .91,  .8} Llama-3.1-8B-Instruct & 32.4  & 1.5   & 2.8   & 32.0  & 0.8   & 1.6  \\
    \rowcolor[rgb]{ .988,  .91,  .8} Llama-3.1-70B-Instruct & 35.1  & 69.9  & 46.7  & 30.7  & 61.8  & 41.0  \\
    \rowcolor[rgb]{ .988,  .882,  .8} Llama-3.3-70B-Instruct & 31.0  & 94.1  & 46.7  & 28.2  & 90.5  & 43.0  \\
    \rowcolor[rgb]{ .882,  .949,  .89} Qwen2.5-Math-7B-Instruct & 7.7   & 91.7  & 14.2  & 6.9   & 88.0  & 12.7  \\
    \rowcolor[rgb]{ .882,  .949,  .89} Qwen2.5-Math-72B-Instruct & 19.5  & 97.3  & 32.5  & 19.0  & 96.3  & 31.7  \\
    \rowcolor[rgb]{ .882,  .933,  .933} Qwen2.5-Coder-7B-Instruct & 2.1   & 99.1  & 4.1   & 0.9   & 98.3  & 1.8  \\
    \rowcolor[rgb]{ .882,  .933,  .933} Qwen2.5-Coder-14B-Instruct & 20.7  & 94.1  & 34.0  & 15.9  & 94.2  & 27.3  \\
    \rowcolor[rgb]{ .882,  .933,  .933} Qwen2.5-Coder-32B-Instruct & 33.4  & 91.2  & 48.9  & 31.5  & 87.6  & 46.3  \\
    \rowcolor[rgb]{ .882,  .922,  .969} Qwen2-7B-Instruct & 22.4  & 10.9  & 14.7  & 20.0  & 8.7   & 12.1  \\
    \rowcolor[rgb]{ .882,  .922,  .969} Qwen2-72B-Instruct & 34.0  & 55.2  & 42.1  & 32.3  & 53.1  & 40.2  \\
    \rowcolor[rgb]{ .89,  .89,  .969} Qwen2.5-7B-Instruct & 26.5  & 33.9  & 29.7  & 26.2  & 28.6  & 27.4  \\
    \rowcolor[rgb]{ .89,  .89,  .969} Qwen2.5-14B-Instruct & 31.5  & 78.8  & 45.0  & 28.3  & 76.3  & 41.3  \\
    \rowcolor[rgb]{ .89,  .89,  .969} Qwen2.5-32B-Instruct & 25.3  & 95.9  & 40.0  & 24.1  & 92.5  & 38.3  \\
    \rowcolor[rgb]{ .89,  .89,  .969} Qwen2.5-72B-Instruct & 38.7  & 92.6  & 54.6  & 36.6  & 90.9  & 52.2  \\
    \rowcolor[rgb]{ .949,  .89,  .988} \textbf{QwQ-32B-Preview} & 61.4  & 54.6  & \textbf{57.8} & 55.7  & 68.0  & \textbf{61.3} \\
    \midrule
    \multicolumn{7}{c}{\textit{Proprietary language models, prompted as \textbf{Critic Models}}} \\
    \midrule
    \rowcolor[rgb]{ .906,  .902,  .902} GPT-4o-0806 & 45.8  & 58.4  & 51.4  & 45.2  & 65.6  & 53.5  \\
    \rowcolor[rgb]{ .816,  .808,  .808} o1-mini & 80.2  & 95.6  & 87.2  & 74.8  & 91.7  & 82.4  \\
    \bottomrule
    \end{tabular}%
  }
  \label{tab:results1_breakdown}%
\end{table}%

\begin{table}[!ht]
  \centering
  \caption{
  For the two PRMs from \citet{skywork-prm}, we additionally adjust the threshold (\S~\ref{subsec:setup}) as the one leading to the highest F1 score on each subset (i.e., each subset adopts a respective optimal threshold), which can be viewed as the two PRMs' \textit{upper bound} performance on \textsc{ProcessBench}.
  This table presents the results on the OlympiadBench and Omni-MATH subsets, which are slightly higher than those in Table~\ref{tab:results1_breakdown} that all adopt the threshold selected on the GSM8K subset.
  }
  \vspace{-1mm}
  \scalebox{0.95}{
    \begin{tabular}{lcccccc}
    \toprule
    \multirow{2}[2]{*}{\textbf{Model}} & \multicolumn{3}{c}{\textbf{OlympiadBench}} & \multicolumn{3}{c}{\textbf{Omni-MATH}} \\
    \cmidrule(lr){2-4}    \cmidrule(lr){5-7}
    & \textbf{error} & \textbf{correct} & \textbf{F1} & \textbf{error} & \textbf{correct} & \textbf{F1} \\
    \midrule
    \rowcolor[rgb]{ .988,  .89,  .8} Skywork-PRM-1.5B (respective thresholds) & 15.3  & 47.5  & 23.1  & 14.0  & 58.5  & 22.6  \\
    \rowcolor[rgb]{ .988,  .89,  .8} Skywork-PRM-7B (respective thresholds) & 18.9  & 48.1  & 27.1 & 14.4  & 58.1  & 23.1  \\
    \bottomrule
    \end{tabular}%
  }
  \label{tab:results1_skywork}%
\end{table}%

\begin{table}[!ht]
  \centering
  \caption{
  Breakdown of evaluation results of the open-source language models (prompted as critic models) using \textit{greedy decoding}.
  }
  \vspace{-1mm}
  \scalebox{0.95}{
    \begin{tabular}{lcccccc}
    \toprule
    \multirow{2}[2]{*}{\textbf{Model}} & \multicolumn{3}{c}{\textbf{GSM8K}} & \multicolumn{3}{c}{\textbf{MATH}} \\
    \cmidrule(lr){2-4}\cmidrule(lr){5-7}
    & \textbf{error} & \textbf{correct} & \textbf{F1} & \textbf{error} & \textbf{correct} & \textbf{F1} \\
    \midrule
    \rowcolor[rgb]{ .988,  .933,  .8} Meta-Llama-3-8B-Instruct & 28.5  & 9.3   & 14.1  & 20.9  & 5.7   & 8.9  \\
    \rowcolor[rgb]{ .988,  .933,  .8} Meta-Llama-3-70B-Instruct & 39.6  & 93.8  & 55.7  & 21.9  & 72.2  & 33.6  \\
    \rowcolor[rgb]{ .988,  .91,  .8} Llama-3.1-8B-Instruct & 36.7  & 17.1  & 23.3  & 23.6  & 7.9   & 11.8  \\
    \rowcolor[rgb]{ .988,  .91,  .8} Llama-3.1-70B-Instruct & 57.5  & 77.7  & 66.1  & 37.7  & 53.9  & 44.4  \\
    \rowcolor[rgb]{ .988,  .882,  .8} Llama-3.3-70B-Instruct & 66.2  & 96.9  & 78.6  & 38.4  & 93.1  & 54.4  \\
    \rowcolor[rgb]{ .882,  .949,  .89} Qwen2.5-Math-7B-Instruct & 14.5  & 99.0  & 25.3  & 13.1  & 94.8  & 23.1  \\
    \rowcolor[rgb]{ .882,  .949,  .89} Qwen2.5-Math-72B-Instruct & 45.9  & 96.4  & 62.2  & 34.3  & 94.6  & 50.4  \\
    \rowcolor[rgb]{ .882,  .933,  .933} Qwen2.5-Coder-7B-Instruct & 0.0   & 20.2  & 0.0   & 0.2   & 25.6  & 0.3  \\
    \rowcolor[rgb]{ .882,  .933,  .933} Qwen2.5-Coder-14B-Instruct & 20.3  & 99.0  & 33.7  & 15.2  & 96.1  & 26.2  \\
    \rowcolor[rgb]{ .882,  .933,  .933} Qwen2.5-Coder-32B-Instruct & 50.7  & 93.8  & 65.8  & 39.7  & 88.2  & 54.8  \\
    \rowcolor[rgb]{ .882,  .922,  .969} Qwen2-7B-Instruct & 28.0  & 0.0   & 0.0   & 19.0  & 5.2   & 8.1  \\
    \rowcolor[rgb]{ .882,  .922,  .969} Qwen2-72B-Instruct & 56.5  & 82.4  & 67.0  & 35.5  & 66.7  & 46.4  \\
    \rowcolor[rgb]{ .89,  .89,  .969} Qwen2.5-7B-Instruct & 36.7  & 66.3  & 47.3  & 23.7  & 63.8  & 34.6  \\
    \rowcolor[rgb]{ .89,  .89,  .969} Qwen2.5-14B-Instruct & 47.8  & 93.8  & 63.3  & 40.4  & 86.9  & 55.2  \\
    \rowcolor[rgb]{ .89,  .89,  .969} Qwen2.5-32B-Instruct & 43.0  & 97.9  & 59.8  & 33.3  & 95.6  & 49.4  \\
    \rowcolor[rgb]{ .89,  .89,  .969} Qwen2.5-72B-Instruct & 61.4  & 98.4  & 75.6  & 45.3  & 91.9  & 60.7  \\
    \rowcolor[rgb]{ .949,  .89,  .988} QwQ-32B-Preview & 74.9  & 67.4  & 70.9  & 58.6  & 54.2  & 56.3  \\
    \bottomrule
    \\
    \toprule
    \multirow{2}[2]{*}{\textbf{Model}} & \multicolumn{3}{c}{\textbf{OlympiadBench}} & \multicolumn{3}{c}{\textbf{Omni-MATH}} \\
    \cmidrule(lr){2-4}\cmidrule(lr){5-7}
    & \textbf{error} & \textbf{correct} & \textbf{F1} & \textbf{error} & \textbf{correct} & \textbf{F1} \\
    \midrule
    \rowcolor[rgb]{ .988,  .933,  .8} Meta-Llama-3-8B-Instruct & 17.2  & 0.6   & 1.1   & 17.3  & 4.1   & 6.7  \\
    \rowcolor[rgb]{ .988,  .933,  .8} Meta-Llama-3-70B-Instruct & 20.9  & 41.6  & 27.8  & 20.9  & 50.2  & 29.6  \\
    \rowcolor[rgb]{ .988,  .91,  .8} Llama-3.1-8B-Instruct & 19.1  & 5.6   & 8.7   & 17.1  & 10.0  & 12.6  \\
    \rowcolor[rgb]{ .988,  .91,  .8} Llama-3.1-70B-Instruct & 32.8  & 32.4  & 32.6  & 29.5  & 39.0  & 33.6  \\
    \rowcolor[rgb]{ .988,  .882,  .8} Llama-3.3-70B-Instruct & 30.9  & 90.0  & 46.0  & 27.1  & 86.3  & 41.3  \\
    \rowcolor[rgb]{ .882,  .949,  .89} Qwen2.5-Math-7B-Instruct & 6.4   & 79.1  & 11.8  & 4.7   & 78.0  & 8.9  \\
    \rowcolor[rgb]{ .882,  .949,  .89} Qwen2.5-Math-72B-Instruct & 17.2  & 95.0  & 29.2  & 18.3  & 93.4  & 30.6  \\
    \rowcolor[rgb]{ .882,  .933,  .933} Qwen2.5-Coder-7B-Instruct & 0.0   & 13.3  & 0.0   & 0.0   & 27.8  & 0.0  \\
    \rowcolor[rgb]{ .882,  .933,  .933} Qwen2.5-Coder-14B-Instruct & 9.1   & 95.6  & 16.6  & 6.2   & 95.9  & 11.6  \\
    \rowcolor[rgb]{ .882,  .933,  .933} Qwen2.5-Coder-32B-Instruct & 31.8  & 86.7  & 46.5  & 31.5  & 84.6  & 45.9  \\
    \rowcolor[rgb]{ .882,  .922,  .969} Qwen2-7B-Instruct & 14.1  & 2.9   & 4.9   & 13.7  & 2.9   & 4.8  \\
    \rowcolor[rgb]{ .882,  .922,  .969} Qwen2-72B-Instruct & 33.4  & 48.1  & 39.4  & 30.4  & 48.1  & 37.3  \\
    \rowcolor[rgb]{ .89,  .89,  .969} Qwen2.5-7B-Instruct & 25.4  & 46.0  & 32.7  & 26.1  & 43.6  & 32.6  \\
    \rowcolor[rgb]{ .89,  .89,  .969} Qwen2.5-14B-Instruct & 30.9  & 76.4  & 44.0  & 27.0  & 72.6  & 39.4  \\
    \rowcolor[rgb]{ .89,  .89,  .969} Qwen2.5-32B-Instruct & 22.4  & 90.0  & 35.9  & 22.4  & 87.6  & 35.7  \\
    \rowcolor[rgb]{ .89,  .89,  .969} Qwen2.5-72B-Instruct & 33.7  & 88.5  & 48.9  & 33.7  & 88.4  & 48.8  \\
    \rowcolor[rgb]{ .949,  .89,  .988} QwQ-32B-Preview & 37.8  & 31.9  & 34.6  & 29.5  & 41.9  & 34.6  \\
    \bottomrule
    \end{tabular}%
  }
  \label{tab:results_greedy}%
\end{table}%

\end{document}